\documentclass{article} 
\usepackage{arxiv}

\usepackage{subcaption}
\usepackage{placeins}
\usepackage{mathtools}
\usepackage{amssymb}
\usepackage{amsmath}
\usepackage{amsfonts}   
\usepackage{fancyhdr}
\usepackage{times}
\usepackage{amsthm}
\usepackage{siunitx}
\usepackage{nicefrac}
\usepackage{listings}

\usepackage{todonotes}
\usepackage{algorithm}
\usepackage{algorithmic}

\lstdefinestyle{mystyle}{
    commentstyle=\color{codegreen},
    keywordstyle=\color{magenta},
    backgroundcolor = \color{grayish}
}

\usepackage{booktabs}  
\usepackage{nicefrac}
\usepackage{microtype}
\usepackage{float}
\usepackage{enumitem}
\usepackage{xcolor}

\definecolor{darkblue}{RGB}{0,0,128}
\definecolor{darkgreen}{RGB}{0, 128, 0}
\definecolor{darkred}{RGB}{128, 0, 0}
\definecolor{black}{RGB}{0, 0, 0}
\definecolor{errorcolor}{HTML}{481567}
\definecolor{viridisgreen}{HTML}{55C667}

\usepackage[backend=biber,
style=apa,
maxcitenames=2,
maxbibnames=100, 
language=english,
doi=false,
isbn=false,
url=false,
uniquename=false
]{biblatex} 
\DeclareLanguageMapping{english}{english-apa} 
\usepackage{hyperref}
\usepackage{url}  
\hypersetup{
	colorlinks=true,
	linkcolor=darkblue,
	filecolor=darkblue,
	urlcolor=darkblue,
	citecolor=darkblue,
	pdfauthor={},
	pdftitle={}
}

\title{BayesFlow: Amortized Bayesian Workflows With Neural Networks}

\author{
    Stefan T. Radev\thanks{shared first authorship}\\
    Cluster of Excellence STRUCTURES\\
    Heidelberg University
    \And
    Marvin Schmitt$^*$\\
    Cluster of Excellence SimTech\\
    University of Stuttgart
    \And
    Lukas Schumacher\\
    Institute of Psychology\\
    Heidelberg University
    \And
    Lasse Elsemüller\\
    Institute of Psychology\\
    Heidelberg University
    \And
    Valentin Pratz\\
    Visual Learning Lab\\
    Heidelberg University
    \And
    Yannik Schälte\\
    Life and Medical Sciences Institute\\
    University of Bonn
    \AND
    Ullrich Köthe\thanks{
    shared senior authorship\\
    \hspace*{5.55mm}\texttt{BayesFlow} is hosted at the public GitHub repository \url{www.github.com/stefanradev93/BayesFlow}
    }\\
    Visual Learning Lab\\
    Heidelberg University
    \And
    Paul-Christian Bürkner$^\dagger$\\
    Department of Statistics\\
    TU Dortmund University
}

\definecolor{codegreen}{rgb}{0,0.6,0}
\definecolor{grayish}{rgb}{0.93, 0.93, 0.93}

\begin{document}
\maketitle

\section{Summary}
Modern Bayesian inference involves a mixture of computational techniques for estimating, validating, and drawing conclusions from probabilistic models as part of principled workflows for data analysis \autocite{burkner_models_2022, gelman_bayesian_2020, schad2021toward}. Typical problems in Bayesian workflows are the approximation of intractable posterior distributions for diverse model types and the comparison of competing models of the same process in terms of their complexity and predictive performance. However, despite their theoretical appeal and utility, the practical execution of Bayesian workflows is often limited by computational bottlenecks: Obtaining even a single posterior may already take a long time, such that repeated estimation for the purpose of model validation or calibration becomes completely infeasible.

\texttt{BayesFlow} provides a framework for \emph{simulation-based} training of established neural network architectures, such as transformers \autocite{vaswani2017attention} and normalizing flows \autocite{papamakarios2021normalizing}, for \emph{amortized} data compression and inference. \emph{Amortized Bayesian inference} (ABI), as implemented in \texttt{BayesFlow}, enables users to train custom neural networks on model simulations and re-use these networks for any subsequent application of the models. Since the trained networks can perform inference almost instantaneously (typically well below one second), the upfront neural network training is quickly amortized. For instance, amortized inference allows us to test a model's ability to recover its parameters \autocite{schad2021toward} or assess its simulation-based calibration \autocite{talts2018, sailynoja2022graphical} for different data set sizes in a matter of seconds, even though this may require the estimation of thousands of posterior distributions. \texttt{BayesFlow} offers a user-friendly API, which encapsulates the details of neural network architectures and training procedures that are less relevant for the practitioner and provides robust default implementations that work well across many applications. At the same time, \texttt{BayesFlow} implements a modular software architecture, allowing machine learning scientists to modify every component of the pipeline for custom applications as well as research at the frontier of Bayesian inference.

\begin{figure}[t]
    \centering
    \includegraphics[width=\linewidth]{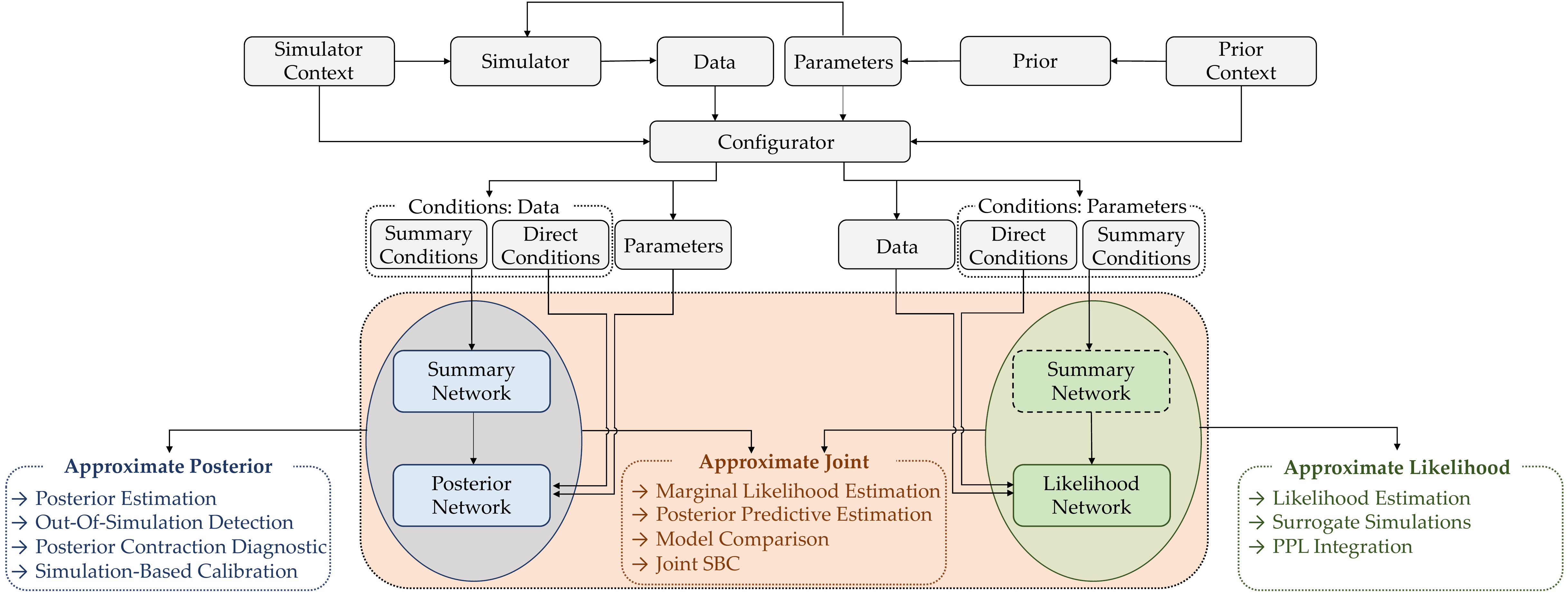}
    \caption{\texttt{BayesFlow} defines a formal workflow for data generation, neural approximation, and model criticism.}
    \label{fig:figure1}
\end{figure}

\section{Statement of Need}

\texttt{BayesFlow} embodies functionality that is specifically designed for building and validating amortized Bayesian workflows with the help of neural networks. \autoref{fig:figure1} outlines a typical workflow in the context of amortized posterior and likelihood estimation. A simulator coupled with a prior defines a generative Bayesian model. The generative model may depend on various (optional) context variates like varying numbers of observations, design matrices, or positional encodings. The generative scope of the model and the range of context variables determine the \emph{scope of amortization}, that is, over which types of data the neural approximator can be applied without re-training. The neural approximators interact with model outputs (parameters, data) and context variates through a configurator. The configurator is responsible for carrying out transformations (e.g., input normalization, double-to-float conversion, etc.) that are not part of the model but may facilitate neural network training and convergence.

\autoref{fig:figure1} also illustrates an example configuration of four neural networks: 1) a summary network to compress simulation outcomes (individual data points, sets, or time series) into informative embeddings; 2) a posterior network to learn an amortized approximate posterior; and 3) another summary network to compress simulation inputs (parameters) into informative embeddings; and 4) a likelihood network to learn an amortized approximate likelihood. \autoref{fig:figure1} depicts the standalone and joint capabilities of the networks when applied in isolation or in tandem. The input conditions for the posterior and likelihood networks are partitioned by the configurator: Complex (``summary'') conditions are processed by the respective summary network into embeddings, while very simple (``direct'') conditions can bypass the summary network and flow straight into the neural approximator.

Currently, the software features four key capabilities for enhancing Bayesian workflows, which have been described in the referenced works:

\begin{enumerate}
    \item \textbf{Amortized posterior estimation:} Train a generative network to efficiently infer full posteriors (i.e., solve the inverse problem) for all existing and future data compatible with a simulation model \autocite{radev2020bayesflow}. 
    \item \textbf{Amortized likelihood estimation:} Train a generative network to efficiently emulate a simulation model (i.e., solve the forward problem) for all possible parameter configurations or interact with external probabilistic programs \autocite{radev2023jana, boelts2022flexible}.
    \item \textbf{Amortized model comparison:} Train a neural classifier to recognize the "best" model in a set of competing candidates \autocite{radev2020evidential, schmitt2022meta, elsemuller2023deep} or combine amortized posterior and likelihood estimation to compute Bayesian evidence and out-of-sample predictive performance \autocite{radev2023jana}.
    \item \textbf{Model misspecification detection:} Ensure that the resulting posteriors are faithful approximations of the otherwise intractable target posterior, even when simulations do not perfectly represent reality \autocite{schmitt2021detecting, radev2023jana}.
\end{enumerate}

\texttt{BayesFlow} has been used for amortized Bayesian inference in various areas of applied research, such as epidemiology \autocite{radev2021outbreakflow}, cognitive modeling \autocite{von2022mental, wieschen2020jumping, sokratous2023ask}, computational psychiatry \autocite{d2020bayesian}, neuroscience \autocite{ghaderi2022general}, particle physics \autocite{bieringer2021measuring}, agent-based econometrics models \autocite{shiono2021estimation}, seismic imaging \autocite{siahkoohi2023reliable}, user behavior \autocite{moon2023amortized}, structural health monitoring \autocite{zeng2023probabilistic}, aerospace \autocite{tsilifis2022inverse} and wind turbine design \autocite{noever2022model}, micro-electro-mechanical systems testing \autocite{heringhaus2022towards}, and fractional Brownian motion \autocite{verdier2022variational}.

The software is built on top of \texttt{TensorFlow} \autocite{abadi2016tensorflow} and thereby enables off-the-shelf support for GPU and TPU acceleration. Furthermore, it can seamlessly interact with TensorFlow Probability \autocite{dillon2017tensorflow} for flexible latent distributions and a variety of joint priors.

\section{Related Software}
When a non-amortized inference procedure does not create a computational bottleneck, approximate Bayesian computation (ABC) might be an appropriate tool. This is the case if a single data set needs to be analyzed, if an infrastructure for parallel computing is readily available, or if repeated re-fits of a model (e.g., cross-validation) are not desired.
A variety of mature Python packages for ABC exist, such as PyMC \autocite{Salvatier2016}, pyABC \autocite{schaelte2022pyabc}, or ELFI \autocite{lintusaari2018elfi}. In contrast to these packages, \texttt{BayesFlow} focuses on amortized inference, but can also interact with ABC samplers (e.g., use BayesFlow to learn informative summary statistics for an ABC analysis).

When it comes to simulation-based inference with neural networks, the \texttt{sbi} toolkit enables both likelihood and posterior estimation using different inference algorithms, such as Neural Posterior Estimation \autocite{papamakarios2021normalizing}, Sequential Neural Posterior Estimation \autocite{greenberg2019automatic} and Sequential Neural Likelihood Estimation \autocite{papamakarios2019sequential}. \texttt{BayesFlow} and \texttt{sbi} can be viewed as complementary toolkits, where \texttt{sbi} implements a variety of different approximators for standard modeling scenarios, while \texttt{BayesFlow} focuses on amortized workflows with user-friendly default settings and optional customization. The \texttt{Swyft} library focuses on Bayesian parameter inference in physics and astronomy. \texttt{Swyft} uses a specific type of simulation-based neural inference technique, namely, Truncated Marginal Neural Ratio Estimation \autocite{miller2021truncated}. This method improves on standard Markov chain Monte Carlo (MCMC) methods for ABC by learning the likelihood-to-evidence ratio with neural density estimators. Finally, the \texttt{Lampe} library provides implementations for a subset of the methods for posterior estimation in the \texttt{sbi} library, aiming to expose all components (e.g., network architectures, optimizers) in order to provide a customizable interface for creating neural approximators. All of these libraries are built on top of \texttt{PyTorch}.

\section{Availability, Development, and Documentation}

\texttt{BayesFlow} is available through PyPI via \texttt{pip install bayesflow}, the development version is available via GitHub. GitHub Actions manage continuous integration through automated code testing and documentation. The documentation is hosted at \url{www.bayesflow.org}. Currently, \texttt{BayesFlow} features seven tutorial notebooks. These notebooks showcase different aspects of the software, ranging from toy examples to applied modeling scenarios, and illustrating both posterior estimation and model comparison workflows.

\section{Acknowledgments}

We thank Ulf Mertens, Marco D'Alessandro, René Bucchia, The-Gia Leo Nguyen, Jonas Arruda, Lea Zimmermann, and Leonhard Volz for contributing to the GitHub repository. STR was funded by the Deutsche Forschungsgemeinschaft (DFG, German Research Foundation) under Germany’s Excellence Strategy - EXC-2181 - 390900948 (the Heidelberg Cluster of Excellence STRUCTURES), MS and PCB were supported by the Cyber Valley Research Fund (grant number: CyVy-RF-2021-16) and the DFG EXC-2075 - 390740016 (the Stuttgart Cluster of Excellence SimTech). LS and LE were supported by a grant from the DFG (GRK 2277) to the research training group Statistical Modeling in Psychology (SMiP). YS acknowledges support from the Joachim Herz Foundation. UK was supported by the Informatics for Life initiative funded by the Klaus Tschira Foundation. YS and UK were supported by the EMUNE project ("Invertierbare Neuronale Netze für ein verbessertes Verständnis von Infektionskrankheiten", BMBF, 031L0293A-D).

\subsubsection*{References}
\printbibliography[heading=none]

@article{wieschen2020jumping,
    year = {2020},
    pages = {120--132},
    title = {Jumping to conclusion? a {L}{\'e}vy flight model of decision making},
    number = {2},
    volume = {16},
    journal = {The Quantitative Methods for Psychology},
    author = {Wieschen, Eva Marie and Voss, Andreas and Radev, Stefan},
}

@article{bieringer2021measuring,
    year = {2021},
    pages = {126},
    title = {Measuring QCD splittings with invertible networks},
    number = {6},
    volume = {10},
    journal = {SciPost Physics},
    author = {Bieringer, Sebastian and Butter, Anja and Heimel, Theo and H{\"o}che, Stefan and K{\"o}the, Ullrich and Plehn, Tilman and Radev, Stefan T},
}

@article{siahkoohi2023reliable,
  title={Reliable amortized variational inference with physics-based latent distribution correction},
  author={Siahkoohi, Ali and Rizzuti, Gabrio and Orozco, Rafael and Herrmann, Felix J},
  journal={Geophysics},
  volume={88},
  number={3},
  pages={R297--R322},
  year={2023},
  publisher={Society of Exploration Geophysicists}
}

@article{greenberg2019automatic,
    year = {2019},
    pages = {2404--2414},
    title = {Automatic posterior transformation for likelihood-free inference},
    volume = {97},
    journal = {International Conference on Machine Learning},
    author = {Greenberg, David and Nonnenmacher, Marcel and Macke, Jakob},
}

@article{radev2020bayesflow,
    doi = {10.1109/TNNLS.2020.3042395},
    year = {2020},
    title = {{BayesFlow}: Learning Complex Stochastic Models With Invertible Neural Networks},
    journal = {IEEE Transactions on Neural Networks and Learning Systems},
    author = {Radev, Stefan T and Mertens, Ulf K and Voss, A and Ardizzone, L and K\"{o}the, U},
}

@article{radev2020evidential,
    year = {2020},
    title = {Amortized {B}ayesian model comparison with evidential deep learning},
    journal = {arXiv preprint},
    author = {Radev, Stefan T and D'Alessandro, Marco and Mertens, Ulf K and Voss, Andreas and K{\"o}the, Ullrich and B{\"u}rkner, Paul-Christian},
}

@article{radev2021outbreakflow,
    year = {2021},
    pages = {e1009472},
    title = {OutbreakFlow: Model-based Bayesian inference of disease outbreak dynamics with invertible neural networks and its application to the COVID-19 pandemics in Germany},
    number = {10},
    volume = {17},
    journal = {PLoS computational biology},
    publisher = {Public Library of Science San Francisco, CA USA},
    author = {Radev, Stefan T and Graw, Frederik and Chen, Simiao and Mutters, Nico T and Eichel, Vanessa M and B{\"a}rnighausen, Till and K{\"o}the, Ullrich},
}

@article{talts2018,
    year = {2018},
    title = {Validating {B}ayesian inference algorithms with simulation-based calibration},
    journal = {arXiv preprint},
    author = {Talts, Sean and Betancourt, Michael and Simpson, Daniel and Vehtari, Aki and Gelman, Andrew},
}

@article{boelts2022flexible,
    year = {2022},
    pages = {e77220},
    title = {Flexible and efficient simulation-based inference for models of decision-making},
    volume = {11},
    journal = {Elife},
    publisher = {eLife Sciences Publications Limited},
    author = {Boelts, Jan and Lueckmann, Jan-Matthis and Gao, Richard and Macke, Jakob H},
}

@article{shiono2021estimation,
    year = {2021},
    pages = {104082},
    title = {Estimation of agent-based models using Bayesian deep learning approach of BayesFlow},
    volume = {125},
    journal = {Journal of Economic Dynamics and Control},
    publisher = {Elsevier},
    author = {Shiono, Takashi},
}

@article{sokratous2023ask,
  title={How to ask twenty questions and win: Machine learning tools for assessing preferences from small samples of willingness-to-pay prices},
  author={Sokratous, Konstantina and Fitch, Anderson K and Kvam, Peter D},
  journal={Journal of Choice Modelling},
  volume={48},
  pages={100418},
  year={2023},
  publisher={Elsevier}
}

@article{verdier2022variational,
  title={Variational inference of fractional Brownian motion with linear computational complexity},
  author={Verdier, Hippolyte and Laurent, Fran{\c{c}}ois and Cass{\'e}, Alhassan and Vestergaard, Christian L and Masson, Jean-Baptiste},
  journal={Physical Review E},
  volume={106},
  number={5},
  pages={055311},
  year={2022},
  publisher={APS}
}

@inproceedings{tsilifis2022inverse,
  title={Inverse design under uncertainty using conditional Normalizing Flows},
  author={Tsilifis, Panagiotis and Ghosh, Sayan and Andreoli, Valeria},
  booktitle={AIAA Scitech 2022 Forum},
  pages={0631},
  year={2022}
}

@article{d2020bayesian,
  title={A Bayesian brain model of adaptive behavior: an application to the Wisconsin Card Sorting Task},
  author={D’Alessandro, Marco and Radev, Stefan T and Voss, Andreas and Lombardi, Luigi},
  journal={PeerJ},
  volume={8},
  pages={e10316},
  year={2020},
  publisher={PeerJ Inc.}
}

@article{heringhaus2022towards,
  title={Towards reliable parameter extraction in MEMS final module testing using bayesian inference},
  author={Heringhaus, Monika E and Zhang, Yi and Zimmermann, Andr{\'e} and Mikelsons, Lars},
  journal={Sensors},
  volume={22},
  number={14},
  pages={5408},
  year={2022},
  publisher={MDPI}
}

@article{noever2022model,
  title={Model updating of wind turbine blade cross sections with invertible neural networks},
  author={Noever-Castelos, Pablo and Ardizzone, Lynton and Balzani, Claudio},
  journal={Wind Energy},
  volume={25},
  number={3},
  pages={573--599},
  year={2022},
  publisher={Wiley Online Library}
}

@article{von2022mental,
    year = {2022},
    pages = {700--708},
    title = {Mental speed is high until age 60 as revealed by analysis of over a million participants},
    number = {5},
    volume = {6},
    journal = {Nature Human Behaviour},
    publisher = {Nature Publishing Group},
    author = {von Krause, Mischa and Radev, Stefan T and Voss, Andreas},
}

@article{burkner_models_2022,
    url = {http://arxiv.org/abs/2209.02439},
    note = {arXiv:2209.02439 [stat]},
    year = {2022},
    title = {Some models are useful, but how do we know which ones? {Towards} a unified {Bayesian} model taxonomy},
    journal = {arXiv preprint},
    urldate = {2023-01-16},
    shorttitle = {Some models are useful, but how do we know which ones?},
    author = {B\"{u}rkner, Paul-Christian and Scholz, Maximilian and Radev, Stefan T.},
}

@article{gelman_bayesian_2020,
    year = {2020},
    title = {Bayesian workflow},
    journal = {arXiv preprint},
    author = {Gelman, Andrew and Vehtari, Aki and Simpson, Daniel and Margossian, Charles C and Carpenter, Bob and Yao, Yuling and Kennedy, Lauren and Gabry, Jonah and B\"{u}rkner, Paul-Christian and Modr\'{a}k, Martin},
}

@article{sailynoja2022graphical,
    year = {2022},
    pages = {32},
    title = {Graphical test for discrete uniformity and its applications in goodness-of-fit evaluation and multiple sample comparison},
    number = {2},
    volume = {32},
    journal = {Statistics and Computing},
    publisher = {Springer},
    author = {S{\"a}ilynoja, Teemu and B{\"u}rkner, Paul-Christian and Vehtari, Aki},
}

@article{elsemuller2023deep,
    year = {2023},
    title = {A Deep Learning Method for Comparing Bayesian Hierarchical Models},
    journal = {arXiv preprint arXiv:2301.11873},
    author = {Elsem{\"u}ller, Lasse and Schnuerch, Martin and B{\"u}rkner, Paul-Christian and Radev, Stefan T},
}

@article{radev2023jana,
    year = {2023},
    title = {JANA: Jointly Amortized Neural Approximation of Complex Bayesian Models},
    journal = {arXiv preprint arXiv:2302.09125},
    author = {Radev, Stefan T and Schmitt, Marvin and Pratz, Valentin and Picchini, Umberto and K{\"o}the, Ullrich and B{\"u}rkner, Paul-Christian},
}

@article{schad2021toward,
    year = {2021},
    pages = {103},
    title = {Toward a principled Bayesian workflow in cognitive science.},
    number = {1},
    volume = {26},
    journal = {Psychological methods},
    publisher = {American Psychological Association},
    author = {Schad, Daniel J and Betancourt, Michael and Vasishth, Shravan},
}

@article{schmitt2021detecting,
    year = {2021},
    pages = {arXiv--2112},
    title = {Detecting Model Misspecification in Amortized Bayesian Inference with Neural Networks},
    journal = {arXiv preprint},
    author = {Schmitt, Marvin and B{\"u}rkner, Paul-Christian and K{\"o}the, Ullrich and Radev, Stefan T},
}

@article{vaswani2017attention,
    year = {2017},
    title = {Attention is all you need},
    volume = {30},
    journal = {Advances in neural information processing systems},
    author = {Vaswani, Ashish and Shazeer, Noam and Parmar, Niki and Uszkoreit, Jakob and Jones, Llion and Gomez, Aidan N and Kaiser, {\L}ukasz and Polosukhin, Illia},
}

@inproceedings{papamakarios2019sequential,
    year = {2019},
    pages = {837--848},
    title = {Sequential neural likelihood: Fast likelihood-free inference with autoregressive flows},
    booktitle = {The 22nd International Conference on Artificial Intelligence and Statistics},
    organization = {PMLR},
    author = {Papamakarios, George and Sterratt, David and Murray, Iain},
}

@inproceedings{abadi2016tensorflow,
    year = {2016},
    pages = {265--283},
    title = {Tensorflow: a system for large-scale machine learning.},
    number = {2016},
    volume = {16},
    booktitle = {Osdi},
    organization = {Savannah, GA, USA},
    author = {Abadi, Mart{\'\i}n and Barham, Paul and Chen, Jianmin and Chen, Zhifeng and Davis, Andy and Dean, Jeffrey and Devin, Matthieu and Ghemawat, Sanjay and Irving, Geoffrey and Isard, Michael and others},
}

@article{ghaderi2022general,
    year = {2022},
    title = {A general integrative neurocognitive modeling framework to jointly describe EEG and decision-making on single trials},
    publisher = {PsyArXiv},
    author = {Ghaderi-Kangavari, Amin and Rad, Jamal Amani and Nunez, Michael D},
}

@article{miller2021truncated,
    year = {2021},
    pages = {129--143},
    title = {Truncated marginal neural ratio estimation},
    volume = {34},
    journal = {Advances in Neural Information Processing Systems},
    author = {Miller, Benjamin K and Cole, Alex and Forr{\'e}, Patrick and Louppe, Gilles and Weniger, Christoph},
}

@article{schmitt2022meta,
    year = {2022},
    title = {Meta-Uncertainty in Bayesian Model Comparison},
    journal = {arXiv preprint arXiv:2210.07278},
    author = {Schmitt, Marvin and Radev, Stefan T and B{\"u}rkner, Paul-Christian},
}

@article{zeng2023probabilistic,
  title={Probabilistic damage detection using a new likelihood-free Bayesian inference method},
  author={Zeng, Jice and Todd, Michael D and Hu, Zhen},
  journal={Journal of Civil Structural Health Monitoring},
  volume={13},
  number={2-3},
  pages={319--341},
  year={2023},
  publisher={Springer}
}

@misc{dillon2017tensorflow,
      title={TensorFlow Distributions}, 
      author={Joshua V. Dillon and Ian Langmore and Dustin Tran and Eugene Brevdo and Srinivas Vasudevan and Dave Moore and Brian Patton and Alex Alemi and Matt Hoffman and Rif A. Saurous},
      year={2017},
      eprint={1711.10604},
      archivePrefix={arXiv},
      primaryClass={cs.LG}
}

@inproceedings{moon2023amortized,
  title={Amortized inference with user simulations},
  author={Moon, Hee-Seung and Oulasvirta, Antti and Lee, Byungjoo},
  booktitle={Proceedings of the 2023 CHI Conference on Human Factors in Computing Systems},
  pages={1--20},
  year={2023}
}

@article{papamakarios2021normalizing,
author = {Papamakarios, George and Nalisnick, Eric and Rezende, Danilo Jimenez and Mohamed, Shakir and Lakshminarayanan, Balaji},
title = {Normalizing Flows for Probabilistic Modeling and Inference},
year = {2021},
issue_date = {January 2021},
publisher = {JMLR.org},
volume = {22},
number = {1},
journal = {Journal of Machine Learning Research},
articleno = {57}
}

@article{lintusaari2018elfi,
author  = {Jarno Lintusaari and Henri Vuollekoski and Antti Kangasr{\"a}{\"a}si{\"o} and Kusti Skyt{\'e}n and Marko J{\"a}rvenp{\"a}{\"a} and Pekka Marttinen and Michael U. Gutmann and Aki Vehtari and Jukka Corander and Samuel Kaski},
title   = {ELFI: Engine for Likelihood-Free Inference},
journal = {Journal of Machine Learning Research},
year    = {2018},
volume  = {19},
number  = {16},
pages   = {1-7},
url     = {http://jmlr.org/papers/v19/17-374.html}
}

@article{Salvatier2016,
  doi = {10.7717/peerj-cs.55},
  url = {https://doi.org/10.7717/peerj-cs.55},
  year  = {2016},
  publisher = {{PeerJ}},
  volume = {2},
  pages = {e55},
  author = {John Salvatier and Thomas V. Wiecki and Christopher Fonnesbeck},
  title = {Probabilistic programming in Python using {PyMC}3},
  journal = {{PeerJ} Computer Science}
}

@article{schaelte2022pyabc,
  title = {pyABC: Efficient and robust easy-to-use approximate Bayesian computation},
  author = {Schälte, Yannik and Klinger, Emmanuel and Alamoudi, Emad and Hasenauer, Jan},
  journal = {Journal of Open Source Software},
  publisher = {The Open Journal},
  year = {2022},
  volume = {7},
  number = {74},
  pages = {4304},
  doi = {10.21105/joss.04304},
  url = {https://doi.org/10.21105/joss.04304},
}
\end{document}